\documentclass{article}

\usepackage[final]{neurips_2019}

\usepackage[utf8]{inputenc} 
\usepackage[T1]{fontenc}    
\usepackage{url}            
\usepackage{booktabs}       
\usepackage{amsfonts}       
\usepackage{nicefrac}       
\usepackage{microtype}      
\usepackage{graphicx}
\usepackage[colorlinks = true,
            linkcolor = blue,
            urlcolor  = blue,
            citecolor = blue,
            anchorcolor = blue]{hyperref}
            
\title{Transform the Set:  Memory Attentive Generation of Guided and Unguided Image Collages}

\author{
 Nikolay Jetchev\\
  \footnotesize{Zalando Research}\\
 \footnotesize{ Berlin, Germany} \\
  \footnotesize{\texttt{nikolay.jetchev@zalando.de}}\\
  \And
  Urs Bergmann\\
   \footnotesize{Zalando Research}\\
   \footnotesize{Berlin, Germany}  \\
   \footnotesize{\texttt{urs.bergmann@zalando.de}} \\
  \And
  Gokhan Yildirim\\
   \footnotesize{Zalando Research}\\
   \footnotesize{Berlin, Germany}  \\
   \footnotesize{\texttt{gokhan.yildirim@zalando.de}} \\
}

\begin{document}

\maketitle
\vspace*{-0.8cm}
\begin{abstract}
Cutting and pasting image segments feels intuitive: the choice of source templates gives artists flexibility in recombining existing source material. Formally, this process takes an image set as input and outputs a collage of the set elements. Such selection from sets of source templates does not fit easily in classical convolutional neural models requiring inputs of fixed size.
Inspired by advances in attention and set-input machine learning, we present a novel  architecture that can generate in one forward pass image collages of source templates using set-structured representations. 
This paper has the following contributions:
(i) a novel framework for image generation called Memory Attentive Generation of Image Collages (MAGIC) which gives artists new ways to create digital collages;
(ii) from the machine-learning perspective, we show a novel Generative Adversarial Networks  (GAN) architecture that uses Set-Transformer layers and set-pooling to blend sets of random image samples -- a hybrid non-parametric approach.
Upon publication of the paper, we will release \href{https://github.com/zalandoresearch/magic}{code} allowing artists and  researchers to use and modify MAGIC.

\end{abstract}

 \setlength{\abovecaptionskip}{0pt}
\setlength{\belowcaptionskip}{-1pt} 
\begin{figure}[h]
    \centering
    \includegraphics[height=6.8cm]{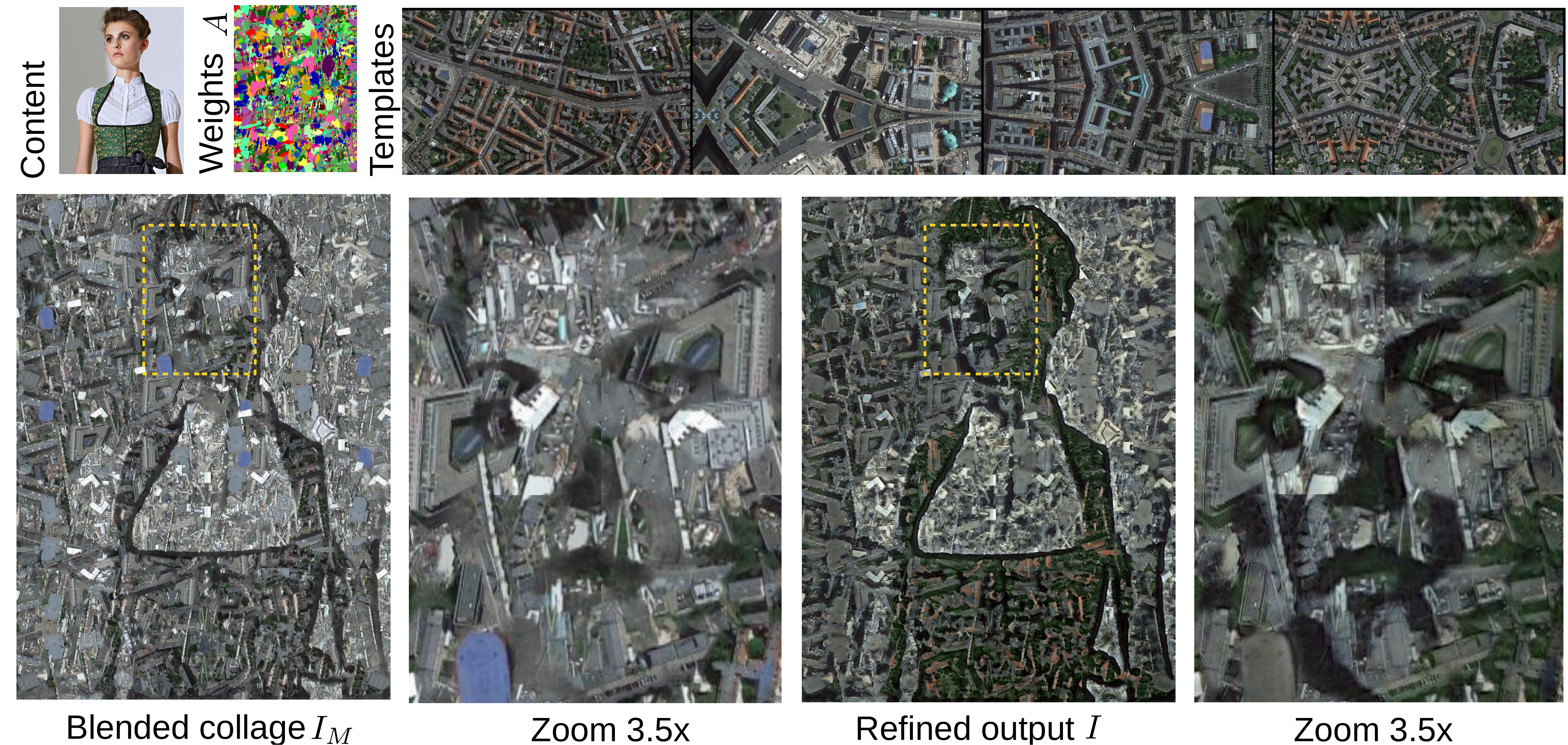}
    \caption{\footnotesize{
    Target content $I_C$ is a human portrait of size 1200x1600 pixels.
    Set $M$ of 50 memory templates is sampled by cropping from few satellite images of Berlin (4 shown).
    MAGIC outputs a collage by predicting blending weights $A$ (visualized with random color for each element). 
    The collage $I_M$ is a convex combination of the memory templates, and output $I$ is further refined by a convolutional network for better perceptual quality. 
    Trained using patches of size 256px. } }
    \label{fig:mosaic}
\end{figure}

 \textbf{Collages} are a classical technique. Throughout ages, artists would stitch pieces of paper for a surreal abstraction or mosaic-like effect to paint a target image \cite{collag}. Modern digital collage methods \cite{JIM} solve an optimization problem to paint a target image by combining a set of memory templates. However, the output has non-overlapping tiles with clear borders. Non-parametric quilting \cite{EfrosQ} combines patches to reconstruct a target image -- "texture transfer". 
However, stitching errors happen and there is no way to ensure perceptually plausible output.
 \cite{liwand16,analogy17} uses neural patch matching (find and copy closest patch in feature space) combined with a content loss, but rely on pre-trained networks and this can negatively impact performance if image distributions differ too much.
As a further drawback, all of \cite{EfrosQ,liwand16,analogy17} are slow due to exhaustive patch search routines, and  optimization is done from scratch for one input memory set and one output image only -- no learning. In contrast, our method MAGIC learns to reason and generalize over the memory sets and their interactions, and so make collages in a single forward pass, amortizing the expensive optimization.

Convolutional networks can be trained to \textbf{represent in their parameters} the statistics of a training image distribution. Learning is adversarial (GANs\cite{Goodfellow14}) or supervised \cite{GatysEB15a}.
They are also used to create smooth mosaics stylizing a target content, using optimization \cite{GatysEB15a, ganosaic} or more efficiently a single forward pass \cite{pix2pix}. However, such stylization is not a collage, since parametric methods do not recombine a set of clearly visible style memory templates.  While GANs trained properly  (big data, big model, long training times) can generate convincing images, this is expensive and data hungry, and can fail to reproduce perfectly all the details of the training  data. In constrast, collaging copies image patches and preserves visual details, without model capacity limitations.  
We propose an architecture that combines collaging (\textbf{non-parametric}: preserve visual details) and GAN approach (\textbf{parametric}: differentiable training, perceptually plausible, fast inference). Our method MAGIC is such a \textbf{hybrid} generative model: using end-to-end learning, it learns how to recombine sampled sets of input memory templates into collages, and it learns how to refine the collages in a smooth way.

Creating a collage from source images is an inherently \textbf{set-structured} problem. The order of the images should not matter -- the method needs to be \textbf{permutation-invariant}. We propose to use the Set Transformer (ST) \cite{SetTransformer} as a way to give the MAGIC model the ability to operate permutation-invariantly on randomly sampled sets of memory templates.
Formally, in a non-parametric collage setting the inputs are \textit{style} images $I_T$, from which we can define a cropping patch distribution $I_t \sim P_T$ of patches of fixed size $H \times W$.
It is used to sample independently $K$ elements ($K$ can also vary randomly) forming the memory templates set $M$. Sampling such sets $M \sim P_T$ is used for input to the generator $G$ -- they replace the usual prior noise distributions for GANs (e.g. Gaussian). The generator first calculates the set of spatial blending weights $A$, and then uses them to generate collage image $I_M$ as a convex combination of the elements of $M$. 
 Please see \hyperref[app:loss]{Appendix I} for details on the generator structure and training loss.

The \textbf{training} of MAGIC optimizes an adversarial loss \cite{Goodfellow14} ensuring that the generated output distribution is perceptually close to the training style distribution $P_T$.
%
Once the model is trained, we can sample a new memory set $M$ and a single forward pass of MAGIC \textbf{quickly infers} the output collage -- this can be useful for  creative exploration of various sampled sets and combinations from them, or smooth stylization of animations and videos. 
We can also view that MAGIC performs relaxed set-structured \textbf{amortized quilting}, since it learns a model to predict blending weights over the whole distribution of memory templates $P_T$. 







 \setlength{\abovecaptionskip}{0pt}
\setlength{\belowcaptionskip}{-5pt} 
\begin{figure}[t]
    \centering
    \includegraphics[height=3.4cm]{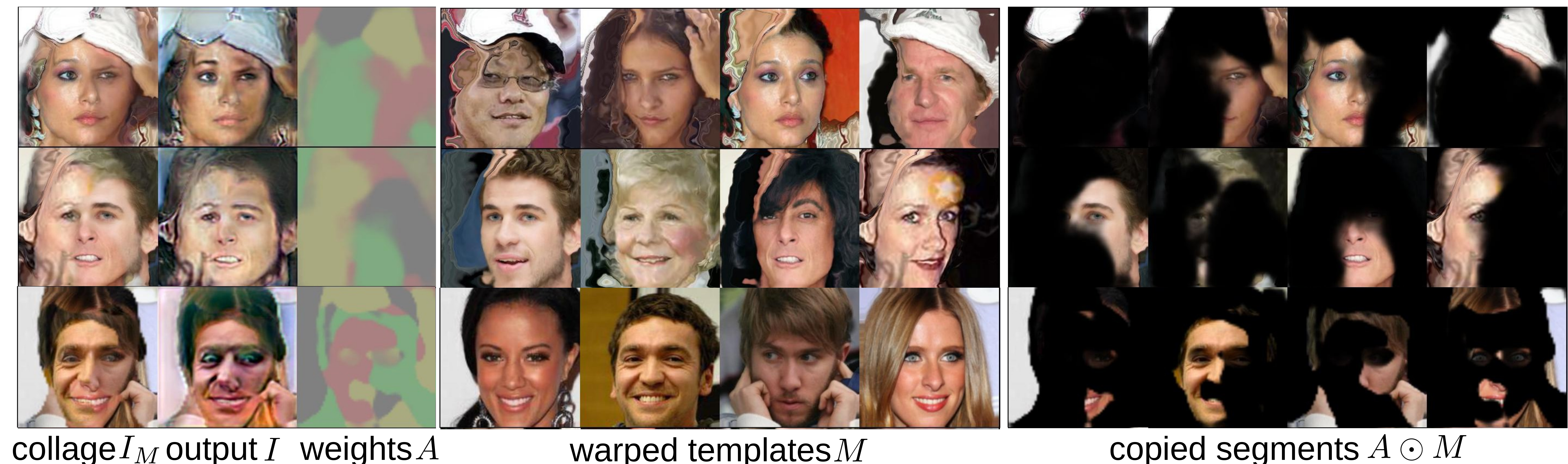}
    \caption{\footnotesize{Using MAGIC for unsupervised generative model learning: given randomly sampled sets of  images $M$ (256x256 pixels), we learn to predict coefficients $A$ to create a blended collage $I_M$, subsequently refined into face $I$.}}
    \label{fig:face}
\end{figure}

\setlength{\belowcaptionskip}{0pt} 

We illustrate how MAGIC works in two different tasks.
First, we can generate seamless collages in a \textbf{content-guided} way (Fig.\ref{fig:mosaic} ). 
We add as additional input  a set of \textit{content} images $I_C$, from which we can define the cropped content patch distribution $I_c \sim P_C$, same size as the style patches. Given patch $I_c$, we would replicate it $K$ times and concatenate it to the templates $M$, i.e. we add 3 extra image channels. 
For training we need content reconstruction loss $\mathcal{L}_{content}$ (\cite{pix2pix,ganosaic}), in addition to the adversarial loss.
The augmented set $M$ informs $G$ how to minimize loss $\mathcal{L}_{content}$ and is informative for the interactions of the template set and the target content.
Once trained on patches, MAGIC inference can be rolled-out on any image size due to the fully convolutional generator. 
The fully convolutional model can create \textbf{very large images} at inference time -- all calls to $G$ can be efficiently split into small chunks seamlessly forming a whole image \cite{SGAN2016}. 

Second, we can learn a collage-based statistical model of data in an \textbf{unsupervised way}, see Fig. \ref{fig:face} for an example on the CelebA dataset \cite{celeba}.
Such generative learning is an interesting alternative to traditional parametric generative models -- MAGIC learns to generate  images recombining elements from the randomly sampled memory template set, "copy and paste", rather then requiring large model capacity to learn the full data statistics. 
In addition to the blending coefficients $A$, the generator can also predict transformation parameters and do warping of each element of the memory set $M$, before blending them into a collage. The integration of \textbf{warping and blending} allows an even more expressive collage model.







In \textbf{related work} FAMOS\cite{FAMOS} also introduced a hybrid  (non-)parametric model to blend templates. However, MAGIC is a more general and advanced method that can cut memory templates in a much finer way, see \hyperref[app:famos]{Appendix III} for detailed comparison.


\pagebreak
 \clearpage

\section*{Acknowledgements}
The authors would like to give special thanks to Kashif Rasul, whose PyTorch expertise greatly aided the project, and to Roland Vollgraf for the many useful generative model discussions.
 
\section*{Ethical Considerations}

The MAGIC algorithm we presented here is a tool for digital collages. It allows artist practitioners to experiment and iterate faster when exploring various artistic choices, e.g. choice of source material memory templates as style or which content image to stylize.
While the new tool amortizes the costs of the collage process -- faster than either manually cutting paper or using traditional optimization based tools -- this tool is inherently meant to be a part of a collaboration between artist and AI machine. In that sense, we would say that our tool is ethical and does not risk radical negative disruption of the artistic landscape (which is a risk when tools completely automate and replace processes). MAGIC is rather a subtle evolution towards finer and faster intelligent control of specific steps involved in the artistic process for a specific artform -- collages and mosaics. Such a tool empowers artists to explore and create more interesting artworks.

\bibliography{bibi}
\setcitestyle{numbers}
\bibliographystyle{plain}

\pagebreak
 \clearpage

\section*{Appendix I: Architecture and loss details}
\label{app:loss}

\begin{figure}[t]
    \centering
    \includegraphics[height=5cm]{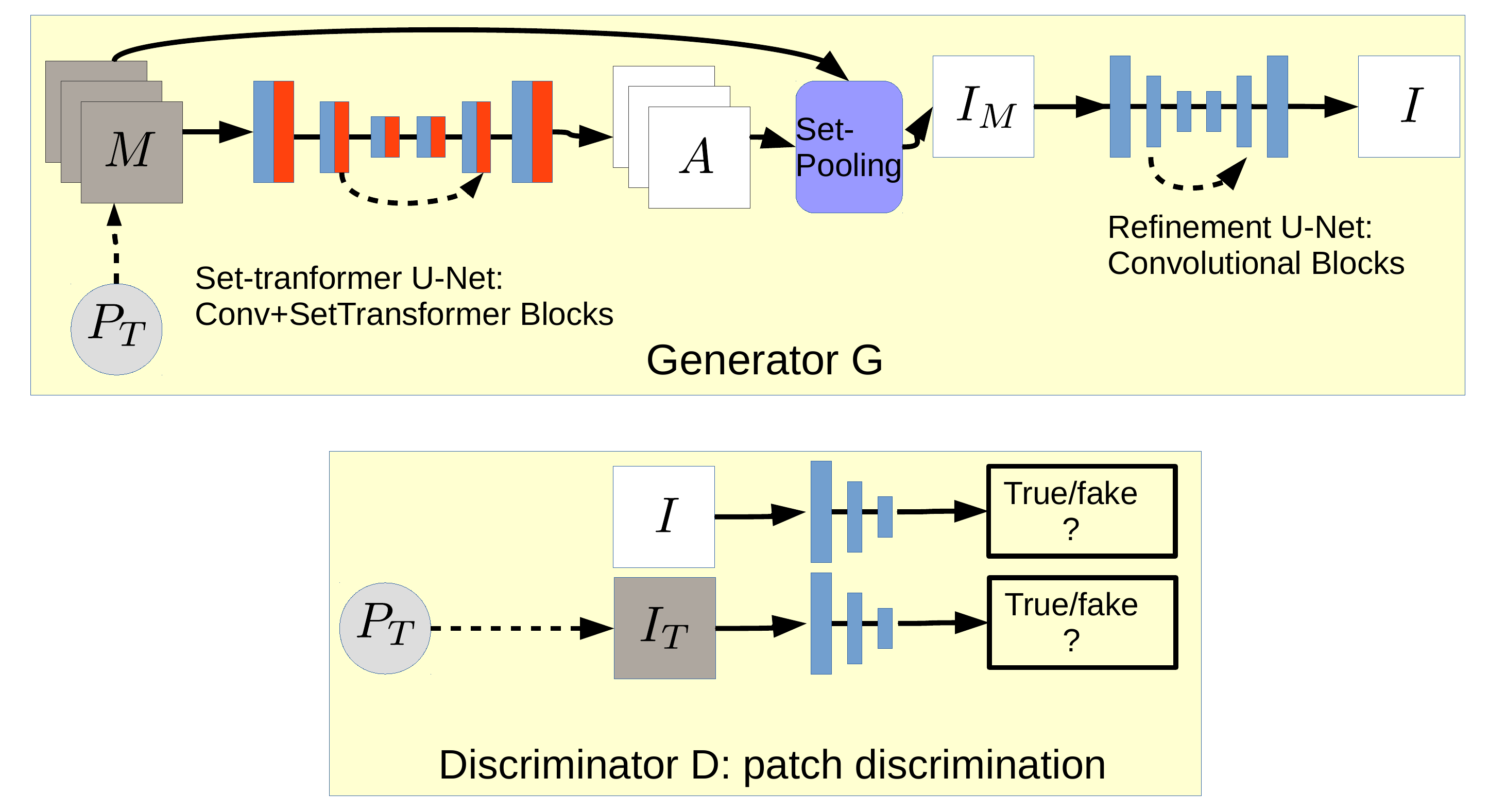}
    \caption{\footnotesize{The generator $G$ has as input a random set $M \sim P_T$ consisting of $K$ images sampled independently -- template memory.
    The architecture of $G$ has  three components.
    (i) a U-Net with ST blocks (see Figure \ref{fig:stc}) can reason about  interactions between set elements and output the blending weights $A$. 
    (ii) The \textbf{permutation-invariant pooling} operation creates the collage image $I_M \in \mathbb{R}^{3 \times H \times W}$ as a convex combination over the $K$ set elements using softmax on $A$.
    (iii)  a purely convolutional U-Net refines $I_M \mapsto I$. The discriminator $D$ distinguishes true patches $I_T \sim P_T$ from the generated patches $I  \sim G(M)$.
    }}
    \label{fig:arch}
\end{figure}

\begin{figure}[t]
    \centering
    \includegraphics[height=5cm]{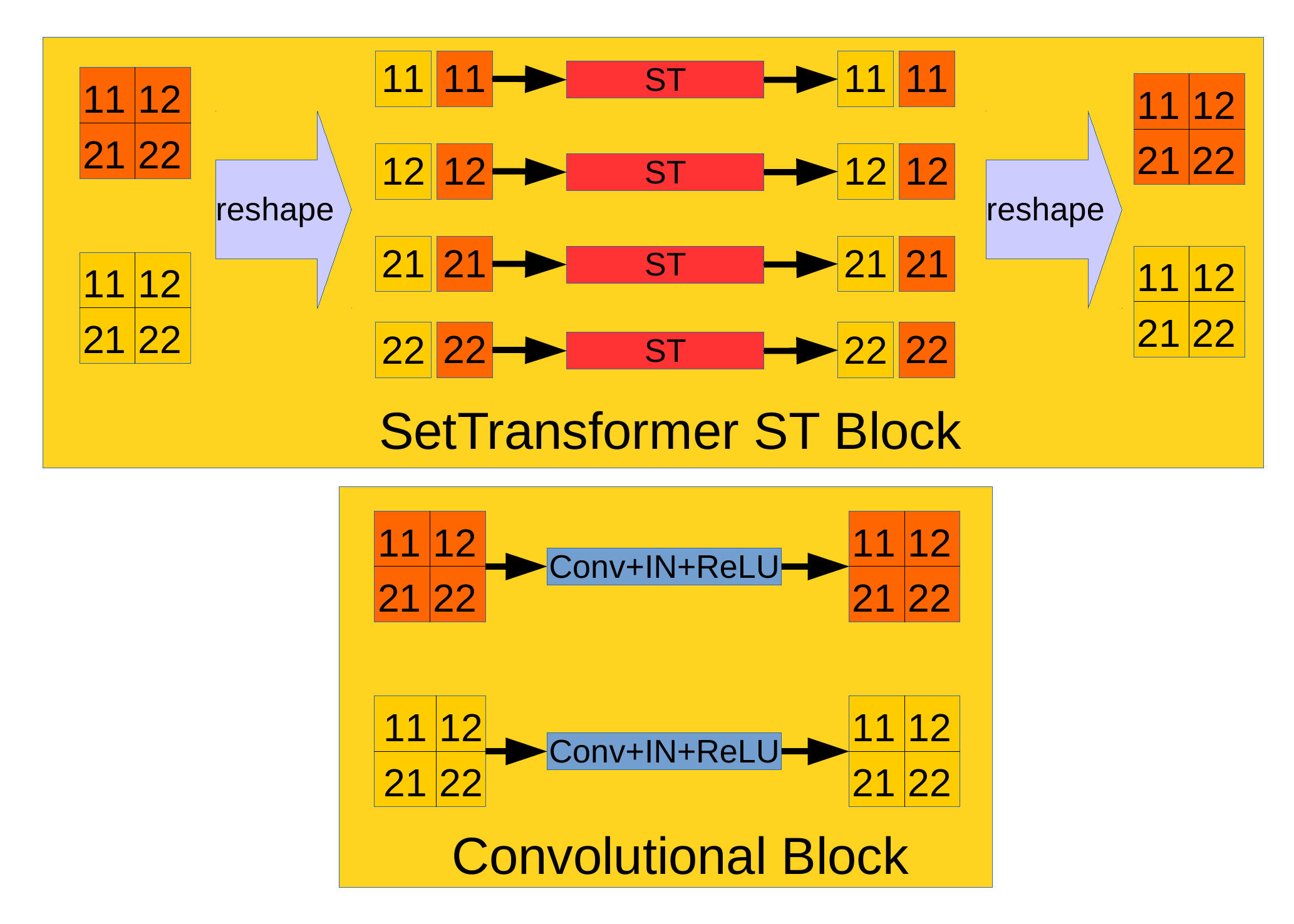}
    \caption{\footnotesize{The structure of the blocks of the U-Nets used inside $G$.
    For illustration, we show how the operation works for $H=W=K=2$. The input tensor is size $X \in \mathbb{R}^{K \times C \times H \times W}$ with $C$ channel feature maps. 
  The set-transformer layer operates on $HW$ number of sets, each of $K$ elements, taken from the same spatial position inside the memory tensor $M$. Each slice $X_{:,:,h,w}$ across the $K$ dimension is a set.
    }}
    \label{fig:stc}
\end{figure}
 
 The structure of the generator $G$ is shown in Figure \ref{fig:arch}.
  The following tensors are used for the generator $G$, (for simplicity we skip minibatch index and write the tensor sizes when having a single instance minibatch)
 
 \begin{itemize}
    \item all spatial patch dimensions are $H \times W$ for training. 
    \item w.l.o.g. we can apply the generator on  spatially larger input tensors $M$ since all convolutional and ST layers can handle such size adjustment.
    \item let $K$ be the number of memory template set elements.
     \item input is the sampled memory set $M \in \mathbb{R}^{K \times 3 \times H \times W}$ 
     \item output of the first ST-U-Net is $A \in \mathbb{R}^{K \times H \times W}$, a set of convex combination weights. By applying softmax on it we ensure that it holds that $\sum_{k=1}^K A_{khw} =1$ .
     \item the collage $I_M \in \mathbb{R}^{3 \times H \times W}$ is the convex combination of  $M$ with $A$ as weights -- this is a form of permutation-invariant set-pooling, since the output does not change if we permute in the first dimension, a slice along the $K$ set elements of set $A$ and set $M$.
     \item the second U-Net takes $I_M$ as input and output $I$, using the parametric architecture of stacked convolutional blocks to correct the collage results and make them more perceptually plausible 
 \end{itemize}

$G$ takes as input randomly sampled memory sets $M$ with $K$ elements (optionally combined with content guidance patches $I_c$).
This is fed to a U-Net (with skip connections as in \cite{pix2pix}). However, in addition to standard convolutional blocks, we also add blocks that can process the set structure of the data  using Set Transformer (ST) layers -- we call this network architecture ST-U-Net. It outputs mixing coefficients $A$ and processes the set elements permutation-equivariantly (the order does not matter) while also taking care of the interactions. See Figure  \ref{fig:stc} for illustration how the set operation is used exactly.
Note that the usual convolutional block is \textbf{permutation-equivariant} by definition, since it works on each element independently.
Afterwards, by applying softmax on $A$, we can calculate a convex combination of the memories from $M$ and produce the collage image $I_M$ with permutation-invariant set pooling.

In addition, we can also optionally do spatial warping on each image the memory set $M$, using a parametrization like the Spatial Transformer \cite{stransf} or directly a full optical flow. For this purpose, we just predict for each set element $k$ its deformation parameters $\theta_k$. We calculate these parameters $\{ \theta_k \}_{k=1}^K$ as output of the ST-U-Net together with $A$,  and apply the warping deformation before the set-pooling.
 
The discriminator $D$ uses a classical patchGAN \cite{liwand16} approach: it should discriminate the sampled training patches from the generated image patches.
The overall loss for training MAGIC (and finding a good generator $G^*$) combines adversarial and content guidance terms (for the content-guidance use case, see \cite{pix2pix,ganosaic}) is the following:

\begin{eqnarray}
 \mathcal{L}_{adv}(G,D)&=E_{I_t \sim P_T}\log D(x)+E_{I_c \sim P_C, M \sim P_T}\log(1-D(G(M,I_c)),\\
\mathcal{L}_{content} &= E_{I_c \sim P_C, M \sim P_T}\|\phi(I_c)-\phi(G(I_c,M))  \|_2,\\
G^* &= \arg\min_G\arg\max_D \mathcal{L}_{adv} + \lambda_{C} \mathcal{L}_{content}.
\end{eqnarray}

\begin{figure}
\centering
\includegraphics[height=4.5cm]{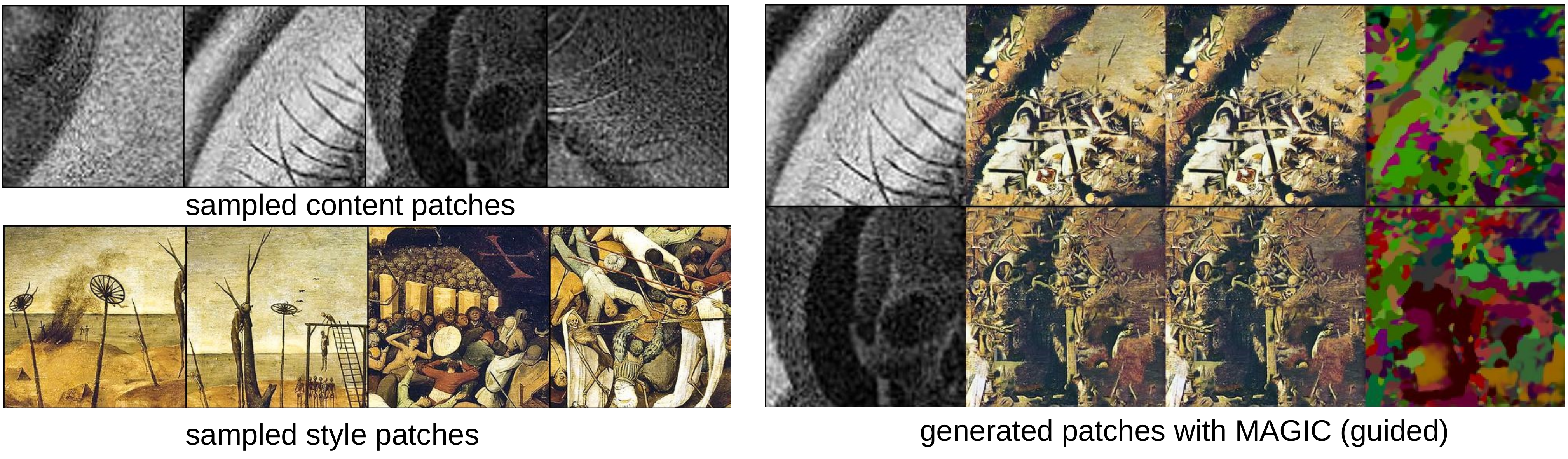} 
\caption{Illustrating the patch-wise training of MAGIC in guided mode.
Training distribution: patches of size $H=W=256x256$ px.}
\label{fig:traininfShow1}
\end{figure}

\begin{figure}
\centering
\includegraphics[height=8.3cm]{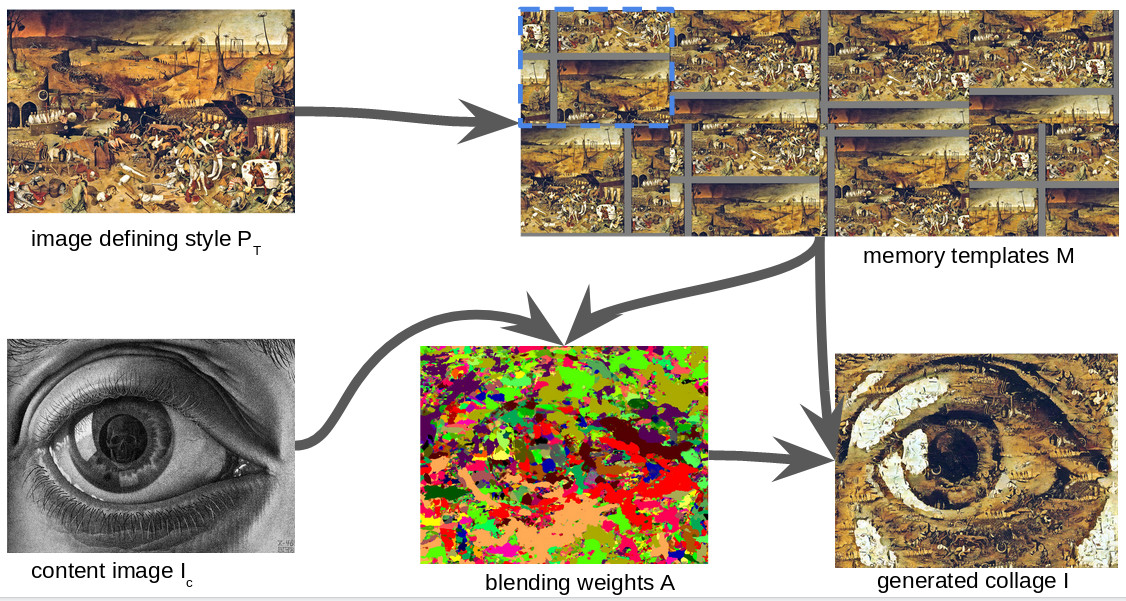}
\caption{Illustrating the inference roll-out of the model.
Sampled collages of size 2100x1600px using seamless stitching of rectangular components. Each component can be cut according to memory constraints}
\label{fig:traininfShow2}
\end{figure}

Figure \ref{fig:traininfShow1} shows visually how patch-based training works for our model, similar to other GAN methods.
Figure \ref{fig:traininfShow2} show how much larger sizes are possible in inference by using convolutional roll-out. We can flexibly do inference on any size image, as long as the memory templates and guided images are cut to the right size. Stitching allows to go beyond the GPU memory limits. E.g. for a size  2000x2000 pixels poster we usually would take 6x6 overlapping grid with overlapping squares of size 384x384 pixels, and trim their edges accordingly (see \citep{SGAN2016}).

\section*{Appendix II: Finer collage control by imposing additional GAN generator constraints}

\begin{figure}[t]
    \centering
    \includegraphics[height=5cm]{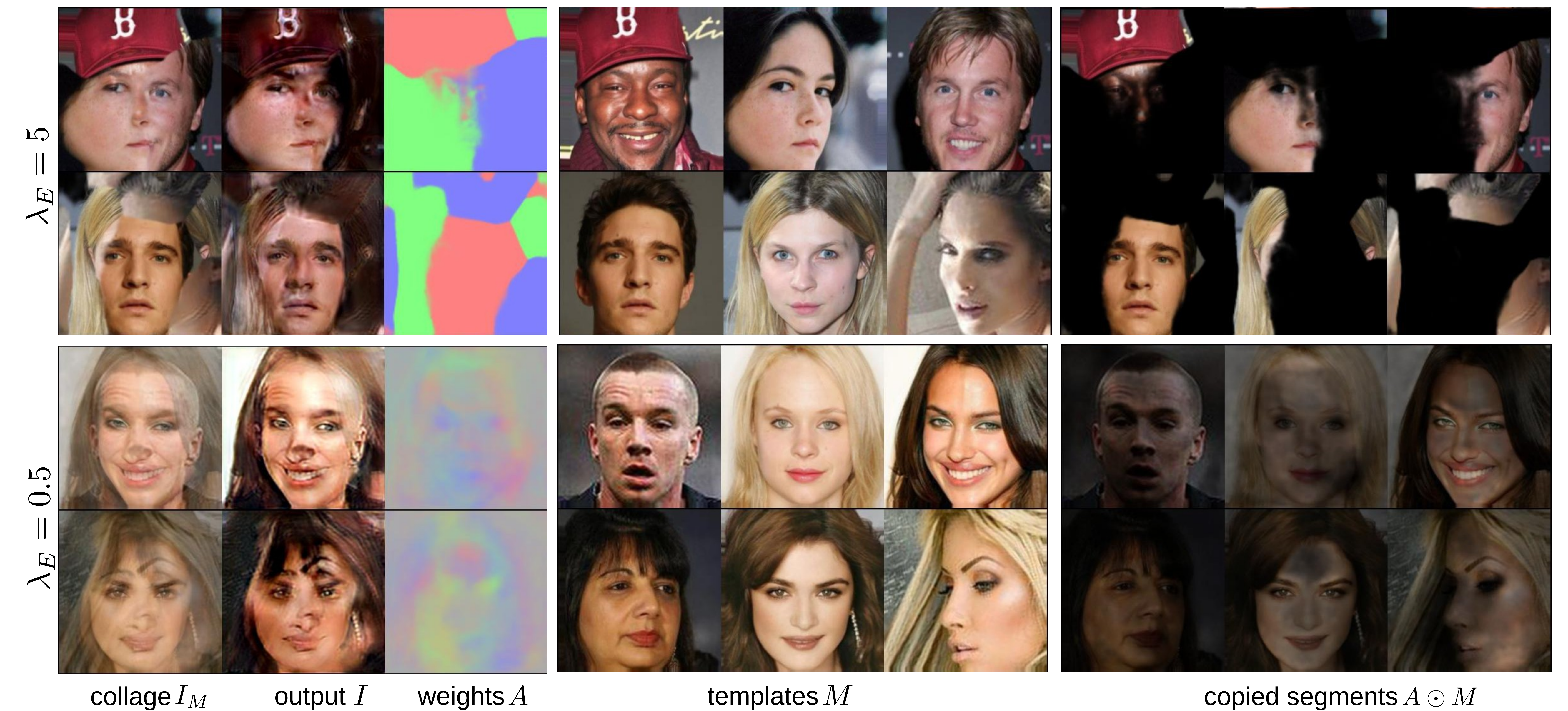}
    \caption{\footnotesize{The effect of entropy regularization of the blending coefficients $A$. Large regularization term weight $\lambda_{E}=5$ leads to solutions with coefficient values close to 0 or 1 -- low entropy, "hard attention" sparse solutions. E.g. the top row shows how we copy the hat, the left and right face parts from 3 different memory templates. Conversely a weaker $\lambda_{E}=0.5$ does not constrain the generator to predict low entropy coefficients $A$ and allows it to relax and soften the blending weights and leads to less sparse collages, blending softly the whole memory images. Example using the CelebA dataset and 256 pixels resolution, with $K=3$ for inference.}}
    \label{fig:entr}
\end{figure}

\begin{figure}[t]
    \centering
    \includegraphics[height=12cm]{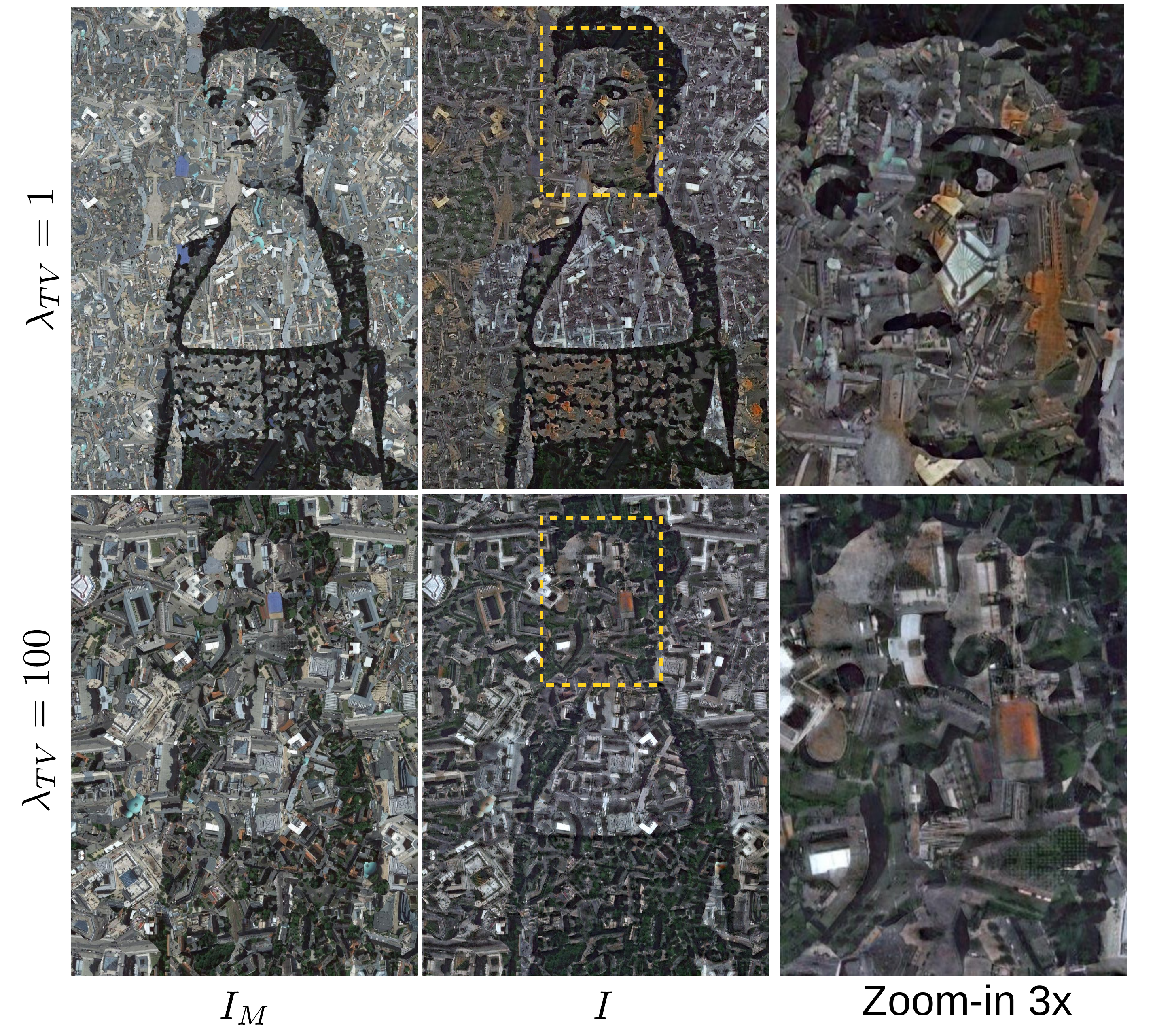}
    \caption{\footnotesize{Example of a guided collage of a human portrait content (size 1200x1600 pixels) and Berlin city fragments used to sample as set of 50 memory templates. We illustrate the effect of TV regularization of the blending coefficients $A$ on the look of the generated collages $I$ and $I_M$.
    A small value $\lambda_{TV}=1$ does not constrain the generator to output smooth $A$ and leads to smaller segment cuts copying smaller details spatially and painting the content more accurately (top row).
    Conversely, large regularization term weight $\lambda_{TV}=100$ leads to solutions with small total variation, implying bigger segments with smoother borders (bottom row). }}
    \label{fig:tv}
\end{figure}

The outputs of the set-transformer U-Net are the convex combination weights $A$.  These determine how the memory templates $M$ are blended (a form of set-pooling). By constraining the generator to output weights $A$ with different statistics, we gain a way to enforce different artistic choices  for the collage generation. 
We experimented with three additional constraints that influence the generator $G$:

\begin{itemize}
\item the entropy $\mathcal{L}_{E}(A)$ determines how sparse the convex combination weights are. Low entropy implies the property of having one memory template be fully copied in a spatial region with weight  $A_k =1$, and others left out with weight $A_k =0$. Conversely, high entropy will be more soft and blend more gently different templates with weight $A_k \approx 0.5$. Please see Figure \ref{fig:entr} for an example how this changes the face blending for unsupervised MAGIC.
\item total variation $\mathcal{L}_{TV}$ determines whether we have bigger segments with small borders, or many small segments that vary spatially. On Figure \ref{fig:tv} we show the effects of such regularization, using as example a large guided mosaic collage.
\item It is desirable to have a collage using a varied selection of the memory template elements $M$. To achieve this, we propose to penalize the spatial size of the largest memory template for the whole spatial region. Formally, we define this term as $\mathcal{L}_M(A) = \max_k  \sum_{hw} A_{khw}$. This term is required especially for the unguided case, where a trivial solution would be to set $A_k =1$ everywhere spatially and copy completely a single memory element. This would fool the discriminator ideally, but is a failure mode for the collage purpose. Our design of $\mathcal{L}_M(A)$ prevents that case easily.
\end{itemize}

By tuning the scalar weights $\lambda$ for  each regularization loss term we can tune the contribution of each regularization term to the total loss, in addition to the adversarial $\mathcal{L}_{adv}$ and content $\mathcal{L}_{content}$ terms we defined already in the previous section:

\begin{eqnarray}
G^* &= \arg\min_G\arg\max_D \mathcal{L}_{adv} + \lambda_{C} \mathcal{L}_{content} + \lambda_{TV} \mathcal{L}_{TV}(A)+\lambda_{E} \mathcal{L}_{E}(A)+\lambda_{M} \mathcal{L}_{M}(A)
\end{eqnarray}

\section*{Appendix III: Detailed comparison with Fully Adversarial Mosaics (FAMOS)}
\label{app:famos}

\begin{figure}[t]
    \centering
    \includegraphics[width=12cm]{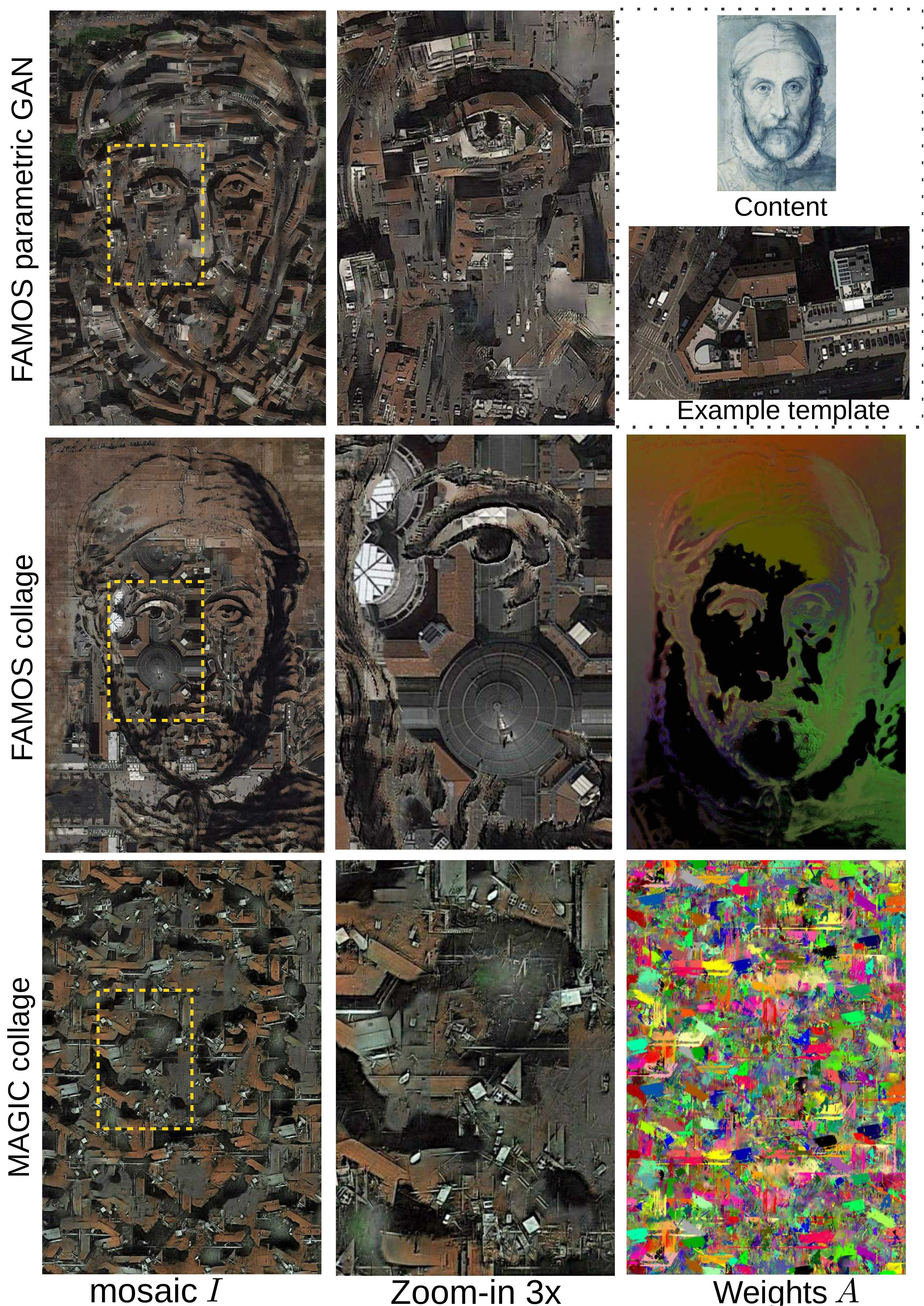}
    \caption{\footnotesize{Illustration of three adversarial guided mosaic stylization approaches, trained on Milan city images as style templates and Archimboldo portrait of size 1200x1800 pixels as content.  (i) A fully convolutional U-Net generator. (ii) A hybrid method that blends from  fixed templates and refines them with another U-Net. (iii) MAGIC collage using the same data. Arguably, MAGIC has the finest control over source material recombination, see text for details.}}
    \label{fig:compf}
\end{figure}

We can compare MAGIC, the method we presented in the current paper, with FAMOS\cite{FAMOS}, another approach proposing a hybrid combination of non-parametric patch copying and fully adversarial end-to-end learning.
The novel ST architecture of MAGIC improves image generation quality and convergence, allowing more flexible cutting of regions of interest from memory templates. Using the set-structure allows to flexibly generalize to randomly sampled sets $M$ -- unlike FAMOS where a predefined ordered tensor $M$ was required, and patch coordinates were "memorized" explicitly.
We can compare visually FAMOS and MAGIC on the guided mosaic task, supported by both models.
We visualize the results of FAMOS, both fully parametric and hybrid memory copying mode, and MAGIC, our novel method. For training we used patches cropped from 4 Milan city satellite images for template distribution, Archimboldo portrait as content Image of size 1200x1800 pixels.

\begin{itemize}
\item The fully parametric convolutional approach (top row) is smooth, but the city image lacks visual details and has some distortion -- the training distribution of Milan city maps is not accurately learned.
\item The hybrid memory template copying and refining approach (middle row) shows a different mosaic result. $K=80$ memory templates were available, fixed for the whole training procedure. A few of the memory templates were copied as background and then the convolutional layers added some more details on top of them, mainly depicting the content image more accurately. However, this collage has quite rough structure: too big segments are copied from the memory templates, see the plot of the mixing coefficients $A$ at the bottom right. While such large memory segment cutting also has a certain charming visual look, it is actually imprecise control of exact patch cutting and placement for the collage.  
\item 
For comparison see the respective MAGIC results (bottom row), with $K=40$ memory templates randomly sampled from a whole distribution. The size of cut and pasted segment is much smaller, and MAGIC can control much better what is copied and pasted where. In addition, despite training with $K=40$ MAGIC can work with different number of set elements, and sample them flexibly -- an advantage of its set transformer generator architecture.
\end{itemize}

 While all 3 tested methods can produce beautiful mosaics with good stylization and content properties, the aesthetic quality of a mosaic is a subjective estimate of the artist or audience. We think that the fine control that MAGIC offers over the placement and cutting of memory templates makes it a worthwhile addition to an AI artist toolbox.

\section*{Appendix IV: Technical details}
We implemented our code using PyTorch 1.0, and ran experiments on a Tesla V100 GPU. Each convolutional block had convolution with kernel 5x5, instance-normalization and ReLU nonlinearity. Typical for U-Nets \cite{pix2pix}, we use downsampling and upsampling to form an hourglass shape. Channels were 48 at the largest spatial resolution and doubled when the spatial resolution was halved.
For the discriminator, we could use much more channels, 128 at the first layer and doubling after every layer.
We also used FP16 precision in order to get lower memory costs and fit larger set sizes $K$ -- note that the complexity of the ST block is square in $K$.
The U-Nets we used had skip connections.

We used for training the usual cross-entropy GAN loss \cite{RadfordMC15}, using minibatches of size $B$, effectively meaning that the sampled memory templates were 5-dimensional tensors $M \in \mathbb{R}^{B \times K \times C \times H \times W}$. Since we use instance normalisation (and not batch normalisation), the batch size $B$ can be chosen flexibly depending on the GPU memory constraints. We trained on 3 V100 GPUs.

For the two experiments we showed in Figures \ref{fig:face} and \ref{fig:mosaic} we had the following settings:
\begin{itemize}
\item guided generation Fig. \ref{fig:mosaic}: raining data distribution for memory templates: randomly cropped 256x256 pixel patches from 6 Berlin city map images (each of resolution 1800x900 pixels).
K=50 memory templates, U-Nets of depth 5. Discriminator depth 6. Batch size $B=6$. For inference of the large output mosaic we can unroll on any size (typically for posters many megapixels can be used), and decide how to split spatially the rendering given the system GPU memory constraints. 
\item unguided generation Fig. \ref{fig:face}: image size 256x256 pixels, same size for training and inference. 
U-Nets of depth 5. Discriminator depth 7. Batchsize $B=12$. Element count  of memory set $K \in \{2,12\}$ sampled randomly at each minibatch iteration.
\end{itemize}

We used the standard ADAM\cite{KingmaB14} optimizer settings as in \cite{RadfordMC15}.
In general, training a MAGIC model is quite fast, orders of magnitude faster than a respective classical parametric GAN -- e.g. around 15 minutes for the guided example, 1 hour for the unguided example. 
Such quick time to adapt MAGIC to a new dataset and allow sampling convincing generated collages is yet another advantage for artistic exploration. This is much faster than what properly training until convergence a parametric GAN model would require.

However, we note that the integration of warping and blending makes the training of MAGIC more difficult. Depending on the deformation model (optical flow, or spatial transformer with various degrees of freedom) more training iterations may be requires. However, warping is not strictly necessary when the training data patches are well aligned for the unguided case, or for the guided case.





\end{document}